\documentclass[a4paper]{cas-sc}

\usepackage[authoryear]{natbib}
\usepackage{threeparttable}    
\usepackage{multirow}
\usepackage{booktabs}  
\usepackage{amssymb}   
\usepackage{amsmath}   

\def\tsc#1{\csdef{#1}{\textsc{\lowercase{#1}}\xspace}}
\tsc{WGM}
\tsc{QE}
\tsc{EP}
\tsc{PMS}
\tsc{BEC}
\tsc{DE}

\begin{document}
\let\WriteBookmarks\relax
\def\floatpagepagefraction{1}
\def\textpagefraction{.001}

\shorttitle{Ego-HOIBench and HGIR}

\shortauthors{K. Deng et~al.}

\title [mode = title]{Egocentric Human-Object Interaction Detection: A New Benchmark and Method}                      



%
\author[]{Kunyuan Deng}[orcid=0009-0005-9502-0200]



\ead{kun-yuan.deng@connect.polyu.hk}


\credit{Conceptualization, Investigation, Formal analysis, Experiment, Visualization, Writing - Original Draft}

\affiliation[]{organization={Department of Electrical and Electronic Engineering, The Hong Kong Polytechnic University},
    addressline={Hung Hom, Kowloon}, 
    city={Hong Kong SAR},
    country={}}

\author[]{Yi Wang}[orcid=0000-0001-8659-4724]
\credit{Conceptualization, Experiment, Supervision, Writing - Review \& Editing}
\ead{yi-eie.wang@polyu.edu.hk}

\author[]{Lap-Pui Chau}[%
    orcid=0000-0003-4932-0593
   ]
\fnmark[*]
\ead{lap-pui.chau@polyu.edu.hk}

\credit{Supervision, Writing - Review \& Editing, Funding acquisition}




\cortext[cor1]{Corresponding author}



\begin{abstract}
Egocentric human-object interaction (Ego-HOI) detection is crucial for intelligent agents to understand and assist human activities from a first-person perspective. However, progress has been hindered by the lack of benchmarks and methods tailored to egocentric challenges such as severe hand-object occlusion. 
In this paper, we introduce the real-world Ego-HOI detection task and the accompanying Ego-HOIBench, a new dataset with over 27K egocentric images and explicit, fine-grained hand-verb-object triplet annotations across 123 categories. Ego-HOIBench covers diverse daily scenarios, object types, and both single- and two-hand interactions, offering a comprehensive testbed for Ego-HOI research.
Benchmarking existing third-person HOI detectors on Ego-HOIBench reveals significant performance gaps, highlighting the need for egocentric-specific solutions. 
To this end, we propose Hand Geometry and Interactivity Refinement (HGIR), a lightweight, plug-and-play scheme that leverages hand pose and geometric cues to enhance interaction representations.
Specifically, HGIR explicitly extracts global hand geometric features from the estimated hand pose proposals, and further refines interaction features through pose-interaction attention, enabling the model to focus on subtle hand-object relationship differences even under severe occlusion. 
HGIR significantly improves Ego-HOI detection performance across multiple baselines, achieving new state-of-the-art results on Ego-HOIBench.
Our dataset and method establish a solid foundation for future research in egocentric vision and human-object interaction understanding. 
Project page: \href{https://dengkunyuan.github.io/EgoHOIBench/}{https://dengkunyuan.github.io/EgoHOIBench/}

\end{abstract}



\begin{keywords}
Egocentric vision\sep 
Human-object interaction\sep 
HOI detection\sep 
Interaction recognition
\end{keywords}

\maketitle

\section{Introduction}
Egocentric human-object interaction (Ego-HOI) detection aims to accurately localize interacting human-object pairs and infer their interaction relationships from the first-person perspective. This task is fundamental for advanced human-centered scene understanding \citep{qiao2025geometric} and has the potential to drive significant progress in a wide range of applications, such as embodied intelligence \citep{kotar2022interactron, nagarajan2019grounded, nagarajan2021shaping}, mixed reality \citep{wang2023holoassist, salvato2022predicting}, surveillance event detection \citep{lee2024error, masuda2020unsupervised}, and visual question answering \citep{di2024grounded, fan2019egovqa, jia2022egotaskqa}. By leveraging egocentric images, Ego-HOI detection systems can provide context-aware guidance and feedback tailored to the user's actions, enabling more intelligent assistance in daily tasks such as cooking, assembly, and navigation. Moreover, robust Ego-HOI detection is essential for advancing embodied AI, as it enables agents to better understand and imitate human behaviors, learn from demonstrations, and execute complex tasks in dynamic real-world environments. 

\begin{figure*}[!t]
\centering
\includegraphics[width=4in]{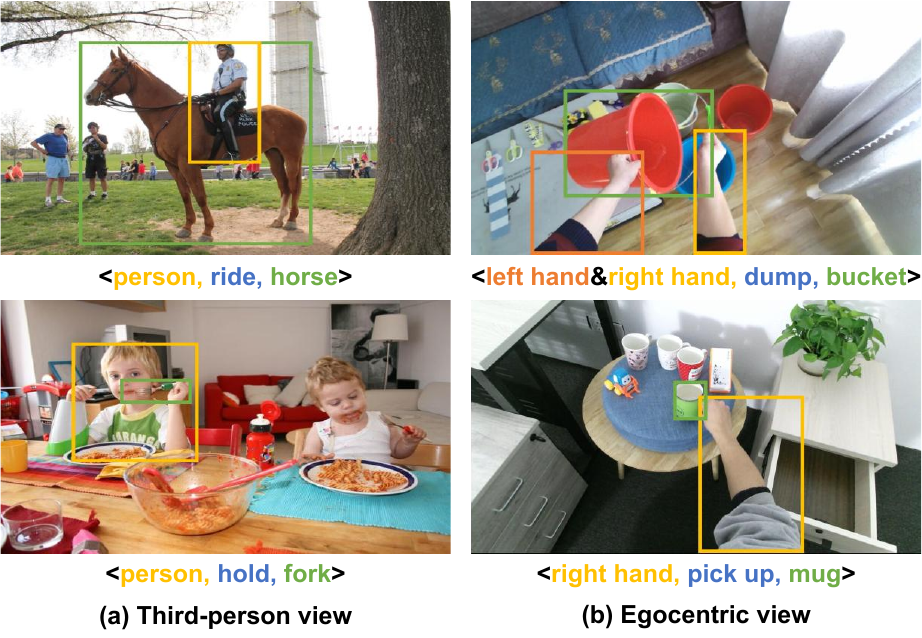}
\caption{Comparison between (a) third-person and (b) egocentric views of human-object interactions. The third-person view provides full-body information and a broader scene context, while the egocentric view offers a close-up, hand-centric imagery with limited context, stronger occlusions, and finer object detail—factors that necessitate different modeling strategies for Ego-HOI. Colored boxes represent different elements of each HOI triplet: human/hand, verb, and object.}
\label{fig1}
\end{figure*}

Despite significant progress in HOI detection for third-person vision \citep{kim2022mstr, yuan2022detecting, hou2022discovering, park2023viplo, yuan2023rlipv2, zheng2023open, ning2023hoiclip, lei2023efficient, chen2023qahoi}, the egocentric perspective remains underexplored. The main barrier is the absence of benchmark datasets with explicit and high-quality annotations for Ego-HOI detection. Existing third-person HOI datasets are unsuitable for egocentric tasks due to substantial domain gaps: third-person images (Fig. \ref{fig1}a) provide holistic body and scene context, while egocentric images (Fig. \ref{fig1}b) focus on close-up hand-object interactions with limited context and frequent occlusions.
Although several egocentric datasets have been introduced, such as Ego4D \citep{grauman2022ego4d} and EPIC-KITCHENS \citep{damen2018scaling, damen2022rescaling}, they primarily target action recognition and lack clear joint annotation of Ego-HOI triplets, i.e., simultaneous connections between human hands, verbs, and objects within each interaction instance. Moreover, most existing datasets suffer from one or more of the following limitations: (1) coverage of only a single scene or activity domain \citep{li2018eye, damen2022rescaling, sener2022assembly101, ragusa2023meccano, ohkawa2023assemblyhands, leonardi2024exploiting}; (2) focus on single-hand interactions \citep{garcia2018first} without distinguishing between left and right hands or supporting two-hand interactions; (3) emphasis on rigid objects, neglecting more complex articulated objects\citep{garcia2018first, kwon2021h2o, sener2022assembly101, ragusa2023meccano, ohkawa2023assemblyhands}. These shortcomings in annotation granularity, interaction diversity, and scenario coverage severely limit the development and evaluation of robust Ego-HOI detection models.

The inherently narrow field of view in egocentric vision significantly increases the likelihood and severity of visual occlusions, as hands and manipulated objects frequently block each other or the scene \citep{su2025care, tekin2019h+, kwon2021h2o, liu2022hoi4d}. This presents a major challenge for interaction recognition, as critical cues for understanding human-object interactions are often partially or fully hidden. Most existing HOI detection methods are developed for third-person vision, where a wide field of view provides abundant contextual and spatial information, reducing the impact of occlusion. When these methods are applied to egocentric settings, their performance degrades due to the loss of contextual cues and the prevalence of mutual occlusions between hands and objects. As a result, Ego-HOI detection requires more robust feature representations that can handle missing or ambiguous visual information. Human pose features, particularly those derived from geometry, have demonstrated greater robustness to partial occlusion due to the structural connectivity of the human skeleton \citep{qiao2022geometric, tu2023agglomerative, rajasegaran2023benefits}. Prior works \citep{gupta2019no, wan2019pose, zheng2020skeleton, qiao2022geometric, li2022transferable, park2023viplo} have explored incorporating pose information to better distinguish subtle interaction differences. 
However, these approaches typically rely on external pose estimators trained on third-person data, which are not optimized for egocentric viewpoints and can hardly capture the fine-grained hand cues critical for Ego-HOI understanding. This highlights the urgent need for methods that can flexibly and accurately capture hand pose information directly within the egocentric context, enabling more robust and reliable interaction recognition.

In view of the above issues, \textbf{1)} we introduce Ego-HOIBench, a comprehensive and rigorously annotated benchmark specifically tailored for Ego-HOI detection. Ego-HOIBench provides explicit \textless human hand, verb, object{\textgreater} triplet annotations for 27,575 real-world images, covering 123 diverse hand-verb-object triplet categories. Each image is thoroughly annotated with bounding boxes and categories for both hands and active objects, as well as their interaction relationships. Our dataset uniquely captures a wide range of egocentric scenarios and objects, including both single-hand and two-hand manipulated interactions, offering an unprecedented testbed for advancing Ego-HOI research. \textbf{2)} We propose Hand Geometry and Interactivity Refinement (HGIR), a novel, lightweight, and plug-and-play scheme that leverages global hand pose and geometric cues to significantly enhance interaction representations. Our approach first estimates multiple candidate hand joint positions from hand features produced by any HOI baseline detector, eliminating the need for external pose estimators. We then construct robust global hand geometric features through a designed selection strategy, and refine interaction representations using pose-interaction attention to generate pose-aware features. Finally, we fuse hand geometric and pose-aware interaction features to achieve superior interaction recognition. Our method is highly efficient, generalizable, and can be seamlessly integrated with existing HOI detectors. The main contributions of our work are summarized as follows:

\begin{itemize}
\item {We introduce the Ego-HOI Detection task and Ego-HOIBench dataset, the first real-world Ego-HOI detection benchmark with explicit joint hand-verb-object triplet annotations, providing a comprehensive and realistic testbed for advancing Ego-HOI research. Besides, we adapt and reimplement four representative third-person HOI detection methods on Ego-HOIBench, aiming to significantly advance the benchmarking works in egocentric interactive recognition and detection research.}
\item {We propose a plug-and-play interaction enhancement scheme, i.e. HGIR, incorporating global hand pose understanding to complement and enhance interaction representations in the egocentric vision. Our approach is lightweight, effective and works seamlessly with off-the-shelf HOI detection methods, enabling substantial performance gains with minimal computational overhead.}
\item {We conduct extensive experiments by integrating our scheme with representative HOI detection baselines, demonstrating consistent and significant performance improvements. Comprehensive ablation studies and analyses further highlight the importance of benchmarking Ego-HOI detection and the effectiveness of our approach.}
\end{itemize}

\section{Related Work}
\subsection{Egocentric Datasets and Benchmarks}
With the rise of wearable devices and smart glasses, such as Apple Vision Pro \citep{wienrich2025immersion} and Microsoft Hololens 2 \citep{balakrishnan2024hololens}, the field of egocentric vision has witnessed a surge in dedicated datasets aimed at understanding human activities from the first-person perspective \citep{damen2018scaling, li2018eye, garcia2018first, kwon2021h2o, jia2022egotaskqa, grauman2022ego4d, liu2022hoi4d, damen2022rescaling, zhang2022fine, zhang2022egobody, sener2022assembly101, ragusa2023meccano, ohkawa2023assemblyhands, leonardi2024exploiting}. Among these, datasets focusing on interaction perception and localization have become particularly important for advancing egocentric understanding. For example, the EPIC-KITCHENS series \citep{damen2018scaling, damen2022rescaling} provides large-scale, densely annotated recordings of unscripted kitchen activities, enabling comprehensive study in realistic environments. EGTEA Gaze+ \citep{li2018eye} enriches egocentric research by offering video, audio, gaze tracking, and detailed action annotations, supporting multi-modal analysis of attention and interaction. Ego4D \citep{grauman2022ego4d} stands out for its unprecedented scale and diversity, introducing a broad suite of benchmarks that cover episodic memory, audio-visual diarization, and social interaction. 

Despite the progress in egocentric vision datasets, there remain significant limitations in providing the joint annotations necessary for Ego-HOI detection models. Most existing datasets focus on action recognition or custom tasks and lack explicit, simultaneous annotations of \textless human hand, verb, object{\textgreater} triplets required for comprehensive Ego-HOI detection \citep{grauman2022ego4d}. For example, many datasets \citep{damen2022rescaling, liu2022hoi4d} annotate all objects interacting with the hand over time but do not specify which object is actively involved in the current interaction, making it difficult to construct precise Ego-HOI labels. The MECCANO dataset \citep{ragusa2023meccano} is one of the few that provides clear active object annotations, but it is limited to toy assembly activities, restricting the generalizability of models trained on it. Furthermore, object and interaction diversity are often lacking: several datasets \citep{garcia2018first, kwon2021h2o, sener2022assembly101, ragusa2023meccano, ohkawa2023assemblyhands} focus mainly on rigid objects and simple interactions, neglecting articulated objects with more complex interactions. Hand configuration diversity is also insufficiently addressed; for instance, FPHA \citep{garcia2018first} only provides 3D joint locations for right hands, while H2O \citep{kwon2021h2o} emphasizes two-hand operations without distinguishing between left and right hands. Even datasets that include both single- and two-hand interactions \citep{li2018eye, garcia2018first} often fail to differentiate hand sides, limiting their utility for detailed Ego-HOI modeling. These gaps in annotation granularity and diversity hinder the development and evaluation of robust Ego-HOI detection models capable of handling real-world complexity.

Table \ref{tab1} summarizes the comparison between our work and existing public datasets. To the best of our knowledge, Ego-HOIBench is the first real image-based dataset to provide explicit and high-quality joint annotations of \textless human hand, verb, object{\textgreater} for Ego-HOI detection, covering a rich diversity of scenarios, objects, and hand configurations, and directly addressing the limitations of prior datasets.

\begin{table*}[t]
   \centering
   \caption{Comparison of Ego-HOIBench and existing egocentric datasets in terms of scenario diversity, annotation modalities, and interaction details. ``Active Object Distinction'' indicates whether the dataset explicitly annotates the object currently involved in the interaction. ``Hand Dist.'' denotes whether left and right hands are distinguished. (*) means can be derived from masks. (**) means available only for a subset of frames.}
   \label{tab1}
  	\resizebox{\linewidth}{!}{      
\begin{threeparttable}          
\begin{tabular}{lccccccccccccc}
\toprule
Dataset                  & Scenario         & Modality & Real & \begin{tabular}[c]{@{}c@{}}\#Interactions/\\ Actions\end{tabular} & \#Nouns & \#Verbs & \begin{tabular}[c]{@{}c@{}}Object\\ BBs\end{tabular} & \begin{tabular}[c]{@{}c@{}}Active Object\\Distinction\end{tabular} & \begin{tabular}[c]{@{}c@{}}Arti-\\ culated\end{tabular} & \begin{tabular}[c]{@{}c@{}}Hand\\ BBs \end{tabular} & \begin{tabular}[c]{@{}c@{}}Hand\\ Poses\end{tabular} & \begin{tabular}[c]{@{}c@{}}Hand\\ Dist.\end{tabular} & Year \\ \hline 
\textit{Single Scenario}             &          &       &         &       &                                                                &                                                   &                                                               &                                                        &                                                  &                                                   &                                                       &  \\
EGTEA Gaze+ \citep{li2018eye}             & Kitchens         & RGB & \checkmark    & 106             & 53        & 19                                                          & $\times$                                                    & $\times$                                                              & \checkmark                                                        & *                                                  & $\times$                                                     & $\times$                                                       & 2018 \\

EPIC-KITCHENS-100 \citep{damen2022rescaling}       & Kitchens         & RGB & \checkmark    & 4,053    & 300       & 97                                                               & \checkmark                                                    & $\times$                                                                  & \checkmark                                                        & \checkmark                                                  & $\times$                                                     & \checkmark                                                      & 2020 \\

Assembly101 \citep{sener2022assembly101}             & Toy assembly  & RGB  &  \checkmark  & 1,380       & 90        & 24                                                               & $\times$                                                     & $\times$                                                               & $\times$                                                         & $\times$                                                   & \checkmark                                                    & N/A                                                    & 2022 \\

MECCANO \citep{ragusa2023meccano}                 & Toy assembly  & RGB-D &  \checkmark  & 61             & 20        & 12                                                          & \checkmark                                                    & \checkmark                                                                 & $\times$                                                         & \checkmark                                                  & $\times$                                                     & \checkmark                                                      & 2022 \\ 
AssemblyHands \citep{ohkawa2023assemblyhands}               & Toy assembly  & RGB  & \checkmark &   N/A         & N/A        & 6                                                              &   $\times$                                                 &  $\times$                                                              &   $\times$                                                       &        $\times$                                           &    \checkmark                                               & \checkmark                                               & 2023 \\
EgoISM-HOI \citep{leonardi2024exploiting}               & Industrial-like   & RGB-D  &$\times$\tnote{2} &   N/A        & 19        & 2                                                            & \checkmark                                                & \checkmark                                                               & \checkmark                                                       & \checkmark                                          &  $\times$                                               & \checkmark                                               & 2024 \\

\hline \hline
\textit{Multiple Scenarios}             &          &       &    &     &       &                                                                &                                                   &                                                               &                                                        &                                                   &                                                   &                                                       &  \\
FPHA \citep{garcia2018first}                    & Daily activities & RGB-D  &\checkmark & 45            & 26        & N/A                                                            & $\times$                                                    & N/A                                                               & $\times$                                                         & $\times$                                                & \checkmark                                                    & $\times$                                                       & 2018 \\
H2O  \citep{kwon2021h2o}                    & Daily activities & RGB-D & \checkmark   & 36             & 8         & 11                                                          & \checkmark                                               & N/A                                                                 & $\times$                                                         & $\times$                                                   & \checkmark                                                    & \checkmark                                                    & 2021 \\
Ego4D \citep{grauman2022ego4d}                   & Multi-domain     & RGB  & \checkmark   & N/A             & 87\tnote{1}        & 74\tnote{1}                                                             & \checkmark                                                    & N/A                                                                 & N/A                                                        & **                                                 & $\times$                                                     & \checkmark                                                     & 2022 \\
HOI4D \citep{liu2022hoi4d}                   & Daily activities & RGB-D  & \checkmark  & N/A          & 16        & 22                                                            & *                                                    & $\times$                                                                  & \checkmark                                                        & *                                                  & \checkmark                                                    & \checkmark                                                      & 2022 \\ 
\textbf{Ego-HOIBench (Ours)} & Daily activities & RGB-D & \checkmark  & 123       & 22        & 18                                                                & \checkmark                                                    & \checkmark                                                                 & \checkmark                                                        & \checkmark                                                  & \checkmark                                                    & \checkmark                                                      & 2025 \\ \bottomrule
\end{tabular}

\begin{tablenotes}    
\item[1] This number is obtained from the “Short-Term Object Interaction Anticipation” task.          
\item[2] Most images are synthetic.
\end{tablenotes}            
\end{threeparttable}       
}                 
\end{table*}

\subsection{HOI Detection}
In recent years, HOI detection has attracted widespread research interest. This task aims to gain a fine-grained understanding of human activities by both localizing human-object pairs and inferring their high-level semantic relationships. Existing HOI detection work are generally divided into two- and one-stage methods based on their detection strategies. 
The two-stage methods \citep{wan2019pose, gao2020drg, zheng2020skeleton, zhang2021spatially, yang2021learning, li2022transferable, zhang2022efficient, zhang2022exploring, park2023viplo, zheng2023open, lei2023efficient, zhang2023exploring} first employ a dedicated object detector (e.g., Faster-RCNN \citep{ren2016faster}, DETR \citep{carion2020end}) to generate candidate human-object pairs, and then classify the interactions between these pairs using a separate recognition network. 
This decoupled pipeline allows for the integration of rich contextual cues, such as spatial relationships \citep{yang2021learning, gao2020drg, zhang2022efficient}, language priors \citep{zheng2023open, lei2023efficient}, and human pose features \citep{wan2019pose, zheng2020skeleton, li2022transferable, park2023viplo}, to enhance interaction understanding. Furthermore, several works \citep{gao2020drg, zhang2021spatially, hong2025deep} utilize graph structures for message propagation between detected human and object instances, thereby enhancing the reasoning performance of interactions between these instance nodes. Decoupling the stages enables training solely the interaction recognition network, thereby saving computational resources and improving training efficiency. However, optimizing the two sub-problems separately may result in suboptimal results, and the reliance on the object detectors can introduce error propagation.

In contrast, one-stage methods \citep{kim2020uniondet, kim2021hotr, kim2022mstr, chen2023qahoi, lin2023point, kim2023relational} aim to directly predict HOI triplets from the entire image in a unified framework. Early CNN-based one-stage approaches, such as those using interaction points \citep{liao2020ppdm} or union boxes \citep{kim2020uniondet}, often depend on heuristic designs and require complex post-processing steps like Non-Maximum Suppression, which may limit their scalability and robustness. The advent of DETR \citep{carion2020end} and its Transformer-based architecture has inspired a new generation of end-to-end HOI detectors \citep{chen2023qahoi, kim2022mstr, kim2021hotr, kim2023relational}, which eliminate the need for hand-crafted post-processing and enable joint optimization of detection and interaction recognition. These methods can be further categorized by their architectural design: single-branch \citep{tamura2021qpic, kim2022mstr}, two-branch \citep{kim2021hotr}, and three-branch \citep{kim2023relational}, depending on how they decouple human, object, and interaction prediction. One-stage approaches offer significant advantages in inference efficiency and end-to-end learning, and have demonstrated strong performance on large-scale benchmarks. However, they may require more training data to generalize well and are less flexible in integrating external cues compared to two-stage pipelines. 
 
The choice between one-stage and two-stage HOI detection methods depends on application needs. To benchmark Ego-HOIBench, we adapt and reimplement several representative and influential third-person HOI detectors in the egocentric setting, including two-stage method STIP \citep{zhang2022exploring}, one-stage one-branch method QPIC \citep{tamura2021qpic}, one-stage two-branch method HOTR \citep{kim2021hotr}, and one-stage three-branch method MUREN \citep{kim2023relational}). 

\subsection{Human Pose as HOI Cues}
Human pose information is a powerful cue for understanding human-object interactions, as it encodes both the intent and the fine-grained dynamics of actions. Numerous studies \citep{gupta2019no, wan2019pose, zheng2020skeleton, wu2022mining, qiao2022geometric, li2022transferable, park2023viplo} have demonstrated that incorporating pose features can significantly enhance interaction recognition, especially in challenging scenarios with occlusions or ambiguous visual contexts. For instance, \cite{park2023viplo} proposed a pose-conditioned graph neural network that enriches human node representations with local joint features, leading to more discriminative interaction modeling. Similarly, \cite{qiao2022geometric} showed that geometric features, such as body posture and object position, can robustly supplement visual features, improving performance under partial occlusion. Hierarchical frameworks like \citep{li2022transferable} further exploit both instance-level and part-level body cues to capture the unique characteristics of interactive body parts.
While pose cues are valuable, most existing methods rely on external pose estimators trained on third-person data, which increases complexity and often fails under egocentric occlusion. To address these issues, we embed hand pose estimation directly into the Ego-HOI detection framework, sharing features with the hand detection branch. This unified, end-to-end strategy yields robust, efficient hand geometry cues tailored for egocentric scenarios, improving both accuracy and practicality.

\section{Our Ego-HOIBench Benchmark}
Ego-HOIBench is an egocentric image dataset designed specifically for Ego-HOI detection research. It provides high-quality, frame-level annotations for both hand-object pair detection and interaction recognition. Each hand and object is annotated with a tuple $\left ( {class, bbox} \right )$, where $class$ indicates the hand side (left or right) or the object category, and $bbox$ specifies the bounding box using the coordinates of the top-left and bottom-right corners. For every hand-object pair, the dataset includes the corresponding interaction category. In addition, original hand pose annotations are provided, offering detailed information for analyzing human-object interactions from an egocentric perspective.

\subsection{Generation Steps}
We generate the Ego-HOIBench dataset through the following steps:
given an untrimmed RGB-D video sequence derived from the HOI4D dataset \citep{liu2022hoi4d}, 
we first extract intermediate frames from each action clip, using the annotated start and end timestamps, to ensure that the selected frames capture the key moments of each interaction.
Next, we associate the mask regions in each intermediate frame with their corresponding object categories. According to the definition of the Ego-HOI detection task, we focus only on the active objects present in each frame. 
By analyzing the task information, we restrict the possible active object categories and filter out irrelevant objects. 
To avoid redundant segmentation, we merge different components of the same object (e.g., the left and right blades of scissors, or the body and door of a safe) into a single object instance. 
Subsequently, we convert the mask regions into bounding boxes. 
For each hand-object pair, we combine the bounding boxes and category labels with the corresponding action category to form a complete Ego-HOI triplet annotation, i.e., \textless human hand, verb, object{\textgreater}.
To ensure annotation quality, all automatically generated annotations are manually reviewed and corrected as needed. Independent annotators cross-check the results to guarantee high-quality ground truth. This manual correction process is time-consuming, requiring a total of about twenty person-days. The entire dataset generation process takes approximately one and a half months.
The extracted intermediate frames and their annotations constitute the Ego-HOIBench dataset. We further divide the dataset into training and test sets by splitting frames according to their video identities, ensuring that object entities do not overlap between the two sets. Using an 80\%/20\% split, we obtain 22,088 training frames and 5,487 test frames.

\subsection{Dataset Statistics}
Ego-HOIBench consists of 27,575 RGB-D frames at a resolution of 1920$\times$1080 pixels. The dataset covers 22 noun categories, including 10 rigid objects, 10 articulated objects, and both left and right hands. In total, 58.4K bounding boxes are annotated, with approximately 28K corresponding to objects.
For action annotation, we define 18 verb categories that reflect typical daily activities performed from a first-person perspective. These verbs are: \textit{Grasp, Pick up, Put down, Carry, Push, Pull, Carry (both hands), Open, Close, Reach out, Turn on, Press, Cut with, Cut, Dump, Dump into, Bind with, Bind}. 

\begin{figure}[!t]
\centering
\includegraphics[width=3.5in]{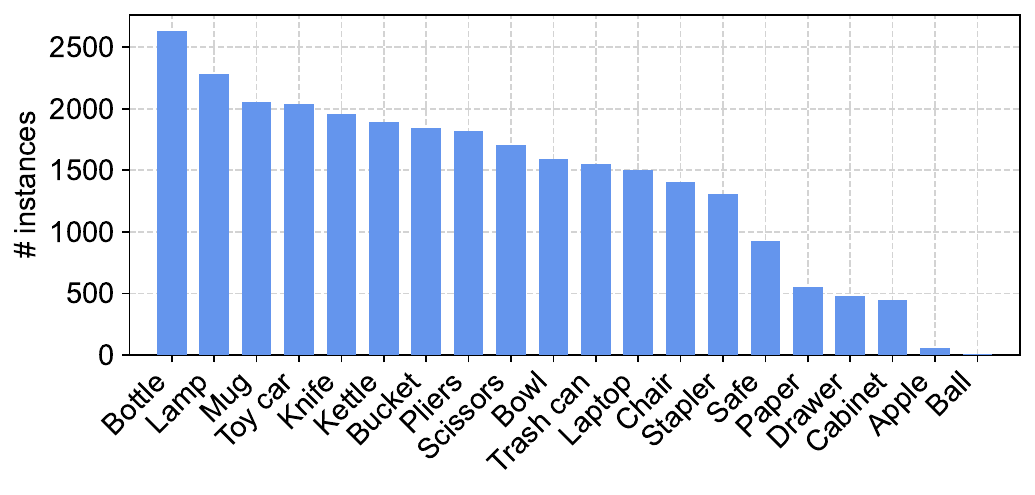}
\includegraphics[width=3.5in]{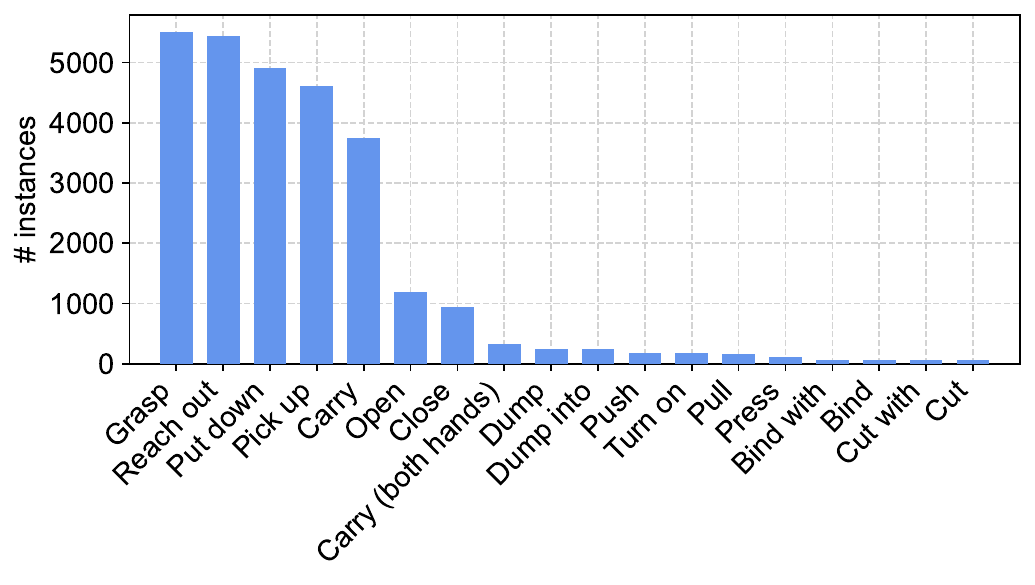}
\caption{Distribution of object (top) and verb (bottom) categories in Ego-HOIBench, sorted by instance count. Most instances are concentrated in a few common categories, while many categories have relatively few examples.}
\label{fig2}
\end{figure}

Among all annotated instances, the majority (91.4\%) involve only the right hand, while 8.2\% involve both hands, and just 0.4\% involve only the left hand. 
Fig.~\ref{fig2} presents the distributions of object and verb categories (defined in Sec.~\ref{task-definition}). The instance number of different object categories varies widely, from as many as 2,630 to as few as 8, and a similar imbalance is observed across verb categories. When considering the full triplet combinations of hands, verbs, and objects, this imbalance becomes even more pronounced, reflecting the natural distribution of HOIs in real-world scenarios. As a result, Ego-HOIBench poses unique and realistic challenges for Ego-HOI detection, making it a valuable and demanding benchmark for practical applications.

\begin{figure*}[!t]
\centering
\vspace{-0.5em}
\includegraphics[width=2.5in]{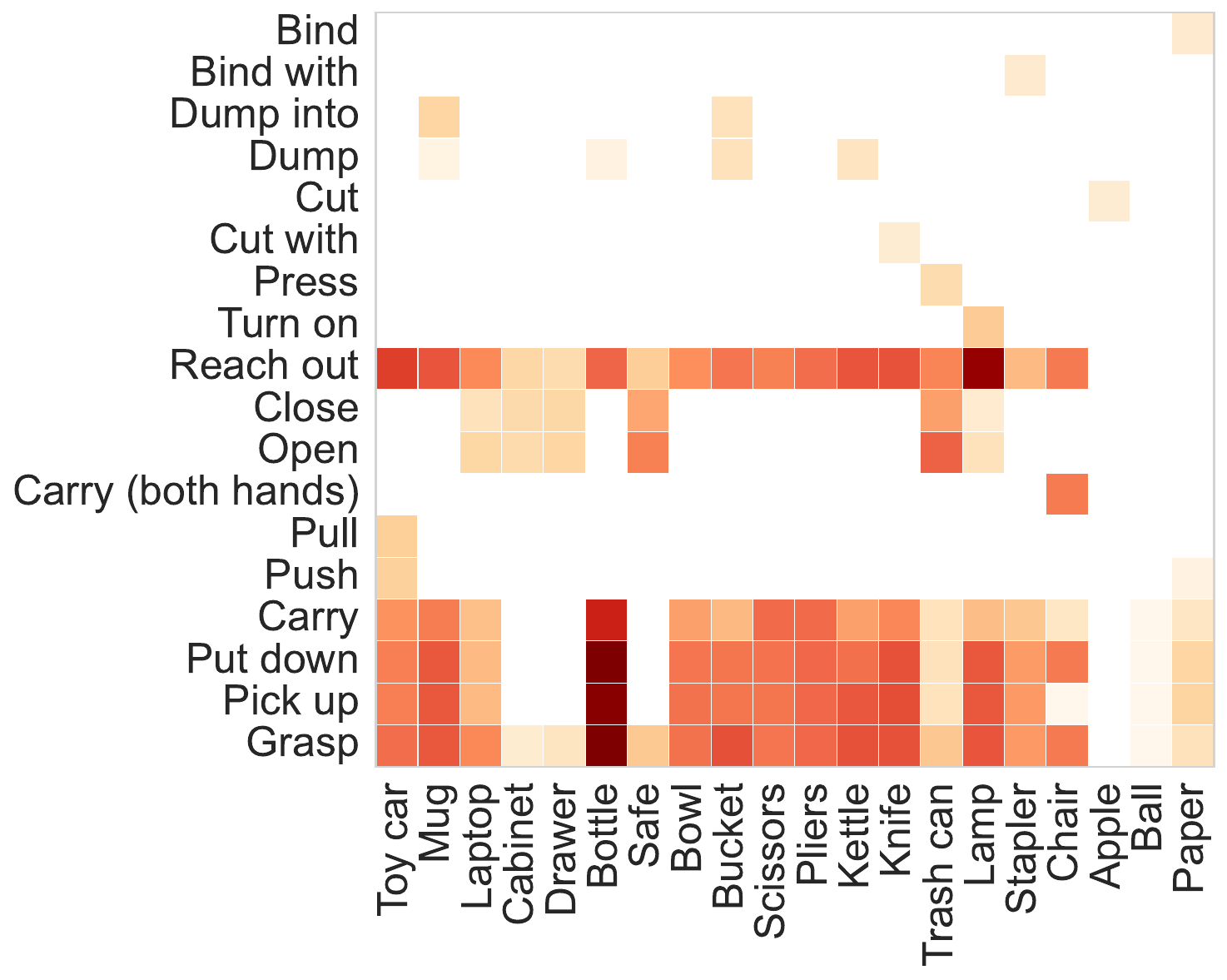}
\caption{Heatmap of verb-object co-occurrence in Ego-HOIBench. Each cell indicates the number of instances for a specific verb-object pair; darker colors represent higher frequencies.}
\label{fig3}
\end{figure*}

Fig.~\ref{fig3} presents a heatmap of verb-object co-occurrence in Ego-HOIBench. Each cell represents the number of instances for a specific verb-object pair, with darker colors indicating higher frequencies. The dataset exhibits a variety of distinctive co-occurrence patterns. 
These co-occurrence statistics reveal that certain verb-object pairs are much more likely to occur than others.
Leveraging this information can help suppress unlikely or negative prediction candidates during model inference, a strategy that aligns with human reasoning and has been widely adopted in recent HOI detection methods.~\citep{gao2020drg, park2023viplo}.

\begin{table*}[t]
\centering
\caption{Number of instances and Ego-HOI triplet categories in the Ego-HOIBench dataset, grouped by object occlusion ratio. The occlusion ratio is defined as the proportion of the object area occluded by hands or other objects.}
\label{tab2}
\resizebox{\linewidth}{!}{      

\begin{tabular}{cccccccccccccc}
\hline
\textbf{}      & \textbf{}         & \textbf{}                  & \multicolumn{5}{c}{\textbf{\#Instances by Occlusion Ratio}}    &  & \multicolumn{5}{c}{\textbf{\#Triplet Categories by Occlusion Ratio}} \\ \cline{4-8} \cline{10-14} 
\textbf{Split} & \textbf{\#Frames} & \textbf{\#Total Ins.} & @0$\sim$0.2 & @0.2$\sim$0.4 & @0.4$\sim$0.6 & @0.6$\sim$0.8 & @0.8$\sim$1 &  & @0$\sim$0.2 & @0.2$\sim$0.4 & @0.4$\sim$0.6 & @0.6$\sim$0.8 & @0.8$\sim$1  \\ \hline
Train          &  22,088                 & 22,481                           & 12,235      & 5,600        & 2,756        &  1,193       &  697     &  &  107       &   105       &  93        &    62      &   42     \\
Test           &    5,487               &  5,567                          & 3,090      & 1,344        &    656     &     325    & 152      &  &  100       &   96       &  75        &   50       & 20       \\ \hline
\end{tabular}}
\end{table*}

Table \ref{tab2} provides a detailed breakdown of Ego-HOI instance counts and unique triplet categories in the Ego-HOIBench training and test sets, grouped by object occlusion ratio. Here, the occlusion ratio is defined as the fraction of an object's bounding box area that is occluded by hands or other objects. Occlusion is common in our dataset: about half of all instances have at least 20\% occlusion, and roughly 20\% of instances are more than 40\% occluded. Higher occlusion ratios make both detection and recognition more difficult, posing significant challenges for model robustness and generalization. Additionally, as occlusion increases, the number of unique triplet categories drops sharply. This is mainly because large objects (such as cabinets or chairs) are usually partially occluded, while small objects (such as staplers or bowls) are more likely to suffer from different degrees of occlusion.

\subsection{Ego-HOI Detection Tasks}\label{task-definition}
Unlike the third-person HOI detection, where the subject is typically treated as a generic ``person'' and subject recognition is ignored~\citep{ning2023hoiclip, dong2022category}, egocentric vision fundamentally requires explicit modeling of the hands. In first-person scenarios, the left and right hands often act independently or collaboratively, each contributing to the interaction. Existing egocentric benchmarks, such as MECCANO~\citep{ragusa2023meccano}, overlook subjects and annotate only verbs and objects, completely omitting explicit hand categories and locations. However, in egocentric HOI, the identity (left/right) and precise position of the hands are critical for correctly interpreting the interaction, as they directly determine the semantics of the action. Therefore, we define Ego-HOI as a \textless hand(s), verb, object\textgreater~triplet, explicitly placing hands at the center of egocentric interaction understanding. This formulation is essential for accurate Ego-HOI detection and enables more fine-grained and robust modeling of real-world egocentric activities.

\begin{figure*}[!t]
\centering
\includegraphics[width=6.4in]{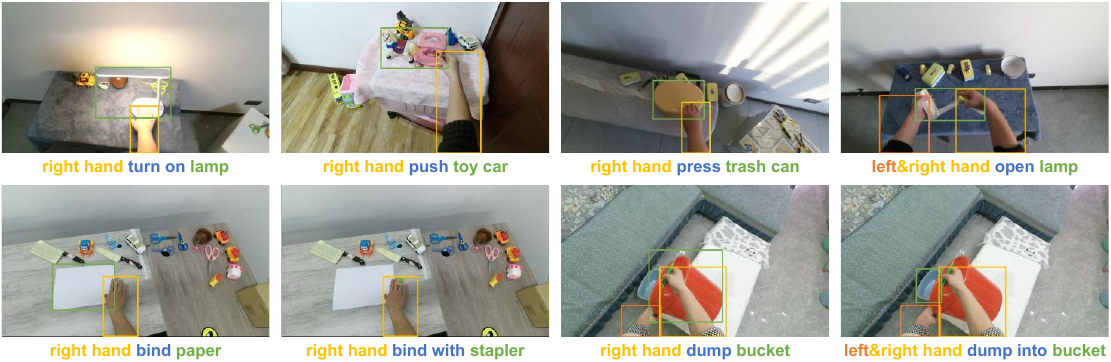}%
\caption{Illustration of \textless hand(s), verb, object\textgreater~triplet annotations for Ego-HOI detection. Each interaction instance is annotated with bounding boxes and categories for the hand(s) and active object, as well as the verb class label.}
\label{fig4}
\end{figure*}

Let $\mathcal{H} = \{ h_{r}, h_{l} \}$ denote the set of hands (right and left), $\mathcal{V} = \{ v_{1}, v_{2}, \dots, v_{m} \}$ the set of verbs, and $\mathcal{O} = \{ o_{1}, o_{2}, \dots, o_{n} \}$ the set of objects, where $m$ and $n$ are the numbers of verb and object categories, respectively. For each interaction instance, the prediction target is defined as:
\begin{equation}
\label{deqn_ex1}
{ehoi}_{ins} = \left \{ \left ( \overline{h_{r}}, \overline{h_{l}} \right ) , v_{i}, o_{j}  \right \}  
\end{equation}
where $\left( \overline{h_{r}}, \overline{h_{l}} \right)$ indicates which hand(s) are involved in the interaction. There are three possible cases: right hand only $\left( h_{r}, \cdot \right)$, left hand only $\left( \cdot, h_{l} \right)$, or both hands $\left( h_{r}, h_{l} \right)$. $v_i \in \mathcal{V}$ is the verb describing the interaction, and $o_j \in \mathcal{O}$ is the object being interacted with. Each instance is annotated with the class labels and bounding boxes for the hand(s) and active object, as well as the verb class label, 
as illustrated in Fig.~\ref{fig4}. In total, we define 123 unique Ego-HOI triplet categories, each consisting of one or two hands, a verb, and an object (e.g., \textit{right-hand cut apple}, \textit{left and right hands dump bucket}). 

\section{Our Method}
In this work, we present a Hand Geometry and Interactivity Refinement (HGIR) scheme that enhances interaction learning in Ego-HOI detection by leveraging global hand pose cues. Our method comprises four components: the hand pose estimation block for obtaining hand pose candidates (see Sec. \ref{sec4.1} for details), the hand geometry extraction block that focuses on exploiting global structural features (see Sec. \ref{sec4.2} for details), the interactivity refinement block by optimizing pose-interaction attention (see Sec. \ref{sec4.3} for details), and the feature aggregation block for fusing complementary geometric and refined interaction features (see Sec. \ref{sec4.4} for details).

\subsection{HGIR Architecture}
\begin{figure*}[!t]
\includegraphics[width=6.5in]{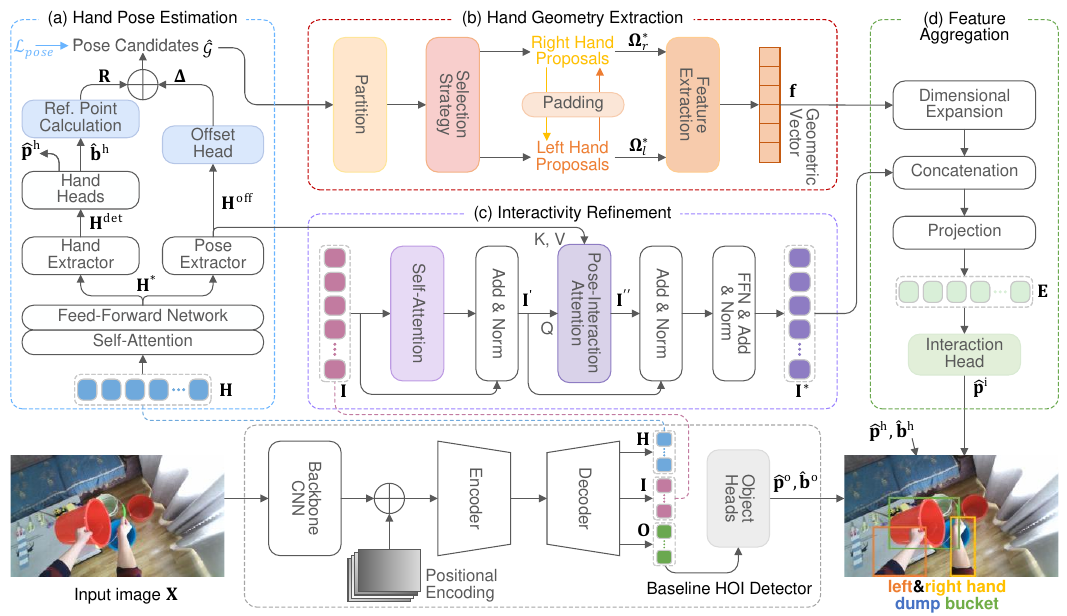}%
\caption{Overview of our framework. Given an input image, a baseline HOI detection method (at the bottom) generates the initial hand ($\mathbf{H}$), object ($\mathbf{O}$), and interaction ($\mathbf{I}$) features. (a) Within our HGIR scheme (at the top), a set of pose candidates ($\hat{\mathcal{G }}$) is first estimated based on $\mathbf{H}$ (see Sec. \ref{sec4.1}). (b) Top K pairs of hand proposals are then selected, and their geometric features ($\mathbf{f}$) are further extracted to reveal the dynamic structural properties of hands in interactions (see Sec. \ref{sec4.2}). (c) Simultaneously, the hand pose offset-specific prompts ($\mathbf{H} ^{\mathrm{off} }$) are incorporated to enrich the interaction representations using the pose-interaction attention mechanism (see Sec. \ref{sec4.3}). (d) Finally, the hand geometric features and refined pose-aware interaction features ($\mathbf{{I}^{\ast} }$) are aggregated to obtain enhanced interaction embedding ($\mathbf{E}$) for interaction recognition (see Sec. \ref{sec4.4}). Our scheme is dedicated to interactivity learning and can be integrated with baseline HOI methods that provide interaction and hand features.}
\label{fig5}
\end{figure*}

Our HGIR scheme is straightforward yet robust and can be easily integrated with various baseline HOI detection methods,
yielding appealing results in the Ego-HOI detection task. The overall architecture of our method is shown in Fig. \ref{fig5}. 
Given an input RGB image $\mathbf{X}  \in \mathbb{R}^{H\times W\times 3}$, we employ the original baseline HOI detection method to obtain the hand features $\mathbf{H} \in \mathbb{R}^ {N\times d}$, the object features $\mathbf{O} \in \mathbb{R}^ {N\times d}$, and the interaction features $\mathbf{I} \in \mathbb{R}^ {N\times d}$, denoted as $\left ( \mathbf{H}, \mathbf{O}, \mathbf{I} \right ) = Baseline \left ( \mathbf{X}   \right ) $. 
The baseline method can adopt either a unified or decoupled prediction strategy as long as it provides the necessary interaction (i.e., verb) and hand (i.e., subject) representations. 
Multiple hand pose candidates $\hat{\mathcal{G }}   \in \mathbb{R}^ {N\times 2N_g}$ are estimated based on $\mathbf{H} $, where $N_g$ is the number of hand joints. Then, a selection strategy is designed to generate left-hand and right-hand pose proposals, and their geometric features $\mathbf{f} \in \mathbb{R} ^{2K N_g \left ( N_g - 1 \right )} $ are extracted to describe the details of hand structure. Simultaneously, the interactivity refinement block uses the attention mechanism to direct the interaction features focus toward the regional information derived from pose offset prompts $\mathbf{H} ^{\mathrm{off} }$. These two features are fused to obtain the ultimate interaction embedding $\mathbf{E} \in \mathbb{R}^ {N \times d} $ for classification. Overall, our HGIR scheme exploits the synergy of complementary geometric features and refined interaction features to enhance the ability to perceive interaction dynamics.

\subsection{Hand Pose Estimation}\label{sec4.1}
Our pose estimation block embeds the auxiliary task of hand pose estimation into the HOI baseline method, sharing most of the network and weights with the hand detection branch. This strategy minimizes computational overhead and allows for flexible adaptation to different datasets without being restricted by the domain of the external hand pose estimator.

Some HOI baseline methods offer specialized hand features, while others use instance features to uniformly describe both the subject and object. To extract and emphasize hand information more deeply, we apply a consistent Transformer encoder across various baseline methods. This encoder is primarily composed of a self-attention layer and a feed-forward (FFN) layer. Formally, we obtain the advanced hand representations using this encoder, denoted as $\mathbf{H }^{*}  = Encoder\left ( \mathbf{H}  \right ) $, where $\mathbf{H }^{*} $ consists of $N$ vectors $\mathbf{h} _i \in \mathbb{R}^{d}$.

Two lightweight multi-layer perceptrons (MLP) are then used in parallel to extract hand detection-specific features $\mathbf{H} ^{\mathrm{det} } $ and pose offset-specific features $\mathbf{H} ^{\mathrm{off} } $, where $i$-th feature vectors are calculated as $\mathbf{h} _{i}^{\mathrm{det} } = MLP\left ( \mathbf{h} _i \right ) $ and $\mathbf{h} _{i}^{\mathrm{off} } = MLP\left ( \mathbf{h} _i \right ) $, respectively. Here, the main reason for choosing MLPs as feature extractors is to ensure the feature index alignment. This index consistency lays the foundation for the subsequent combination of in-box reference points and pose offsets according to the shared indexes.

\textbf{Reference Point.} Two small FFNs $f_{hc}$, $f_{hb}$ are adopted as prediction heads to obtain the hand classification probabilities $\left \{ \mathbf{\hat{p} } _i^\mathrm{h}  \right \} _{i=1}^{N} $ (i.e., left hand or right hand) and bounding boxes $\left \{ \mathbf{\hat{b} } _i^\mathrm{h}  \right \} _{i=1}^{N} $ of all $N$ tokens, respectively, as follows:
\begin{equation}
\begin{aligned}
& \mathbf{\hat{p} } _{i}^{\mathrm{h} } = \delta  \left ( f_{hc}\left ( \mathbf{h} _i^{\mathrm{det} } \right )  \right ) \in \mathbb{R}^{\left | \mathcal{H} \right |  +1}
 \\ & \mathbf{\hat{b} }_{i}^{\mathrm{h} } = \sigma \left ( f_{hb}\left ( \mathbf{h} _i^{\mathrm{det} } \right )  \right )  \in \mathbb{R}^{4}
\end{aligned}
\label{eq: handhead}
\end{equation}
where $\delta$ and $\sigma$ are the sigmoid and softmax operations, respectively. $\left | \mathcal{H} \right |$ denotes the number of hand category set, and the additional class represents the background class (no object). The predicted category $\hat{c} _i^\mathrm{h } $ and score $\hat{s} _i^\mathrm{h} $ are given by 
$\underset{k}{\arg\max\,} \mathbf{\hat{p}}  _{i,k}^\mathrm{h} $ and $\underset{k}{\max\,} \mathbf{\hat{p}}  _{i,k}^\mathrm{h} $.

Using the predicted $N$ hand bounding boxes, we determine the reference points $\mathbf{R}   = \left \{ \left ( xref_i, yref_i \right ) \right \} ^{N}_{i=1} $. 
After in-depth analysis and experimental verification, we choose the top center point of each bounding box as the reference point, which constrains the positions of hand joint points to the vicinity of the hand, making it easier to obtain accurate estimates of joint positions.

\textbf{Offset.} Using an additional offset head, we predict the offsets of $N_g = 21$ hand joints relative to the corresponding reference point from $\mathbf{R}$ along the $x$ and $y$ axes. Taking the hand offset-specific features as inputs, the $i$-th offset vector predicted by the offset head $f_{\Delta} $ is given by:
\begin{equation}
\mathbf{{\Delta}} _{i} = \sigma \left ( f_{\Delta}\left ( {\mathbf{h} }^{\mathrm{off} }_i  \right )  \right )  \in \mathbb{R}^{2N_g}
\end{equation}
where $\left \{ \left ( {\Delta}_{i, 2k-1}, {\Delta}_{i, 2k} \right ) \mid  k=1,\dots, N_g \right \} $ denotes the $x$-coordinate and $y$-coordinate of the $k$-th joint.

The reference points and offsets of the same indexes are added to obtain a set of hand gesture candidates $\hat{\mathcal{G }}  = \left \{\mathbf{\hat{g}}_i  \mid \mathbf{\hat{g}}_i \in \mathbb{R}^ {2N_g}  \right \}^{N}_{i=1} $, as follows:
\begin{equation}
\left ( \mathbf{\hat{g}}  _{i,2k-1},\mathbf{\hat{g}}  _{i,2k} \right ) = \left ( {xref}_i + \mathbf{{\Delta}} _{i,2k-1}, {yref}_i + \mathbf{{\Delta}} _{i,2k} \right ) 
\end{equation}
where $k \in \left \{ 1, \dots, N_g  \right \} $.

Combining reference points and offsets, rather than directly regressing joint positions, provides two significant advantages. First, it constrains the search space for joint localization to the vicinity of detected hands, which greatly improves accuracy and efficiency compared to unconstrained regression over the entire image. Second, the explicit modeling of pose offsets enables the network to learn fine-grained, context-dependent hand configurations, which can be leveraged as informative prompts in the subsequent interactivity refinement module. This strategy not only enhances the robustness of hand pose estimation under occlusion but also facilitates more precise and discriminative interaction reasoning.

\subsection{Hand Geometry Extraction}\label{sec4.2}
From an egocentric view, the hands can carry out tasks independently or collaboratively. Even though the left and right hands perform different actions, they can still provide valuable complementary information to each other. Therefore, we extract the geometric features of all hands in the image from a global perspective based on the pose estimation results to gain a comprehensive insight into the interaction's semantics.

\textbf{Selection Strategy.} We match the hand pose candidates with predicted hand categories and scores to obtain a set of $\left \{ \left (\mathbf{ \hat{g}}  _i,  \hat{c}_i^\mathrm{h} , \hat{s}_i^\mathrm{h}  \right )  \right \} _{i=1}^N $. This matching process also benefits from the index consistency mentioned before. Based on $\hat{c} _i^\mathrm{h}$, the hand pose candidates are partitioned into two sets $\Omega = \left \{ \Omega_{l}, \Omega_{r} \right \} $, where $\Omega _{l}$ and $\Omega _{r}$ denote the set of predictions whose categories are left hand and right hand, respectively. 

To screen out high-quality hand pose candidates, we preset a threshold $T_{pose}$ and the retained left-hand and right-hand pose candidates are denoted as $\Omega_{l}^{' } = \left \{ \mathbf{\hat{g} } _i \in  \Omega_{l} \mid \hat{s} _i^\mathrm{h} \ge T_{pose} \right \} $ and $\Omega_{r}^{' } = \left \{ \mathbf{\hat{g} } _i \in  \Omega_{r} \mid \hat{s} _i^\mathrm{h} \ge T_{pose} \right \} $, respectively. For each set, we re-rank the candidates based on $\hat{s} _i^\mathrm{h}$ and select the Top $K$ candidates with the highest confidence to constitute the pose proposals.
In the case of fewer than $K$ valid candidates, we use the candidates of the other hand for padding to maintain feature integrity. For example, if the number of valid candidates for the left hand is less than $K$, we will use candidates from the right hand for padding, and vice versa:
\begin{equation}\Omega_{l}^{\ast  } = \Omega_{l}^{' } \cup \left \{ \mathbf{\hat{g}} _i \in \Omega_{r}^{' } \mid i=1,\dots , K -\left | \Omega_{l}^{' } \right |   \right \} 
\end{equation}
\begin{equation}
\Omega_{r}^{\ast } = \Omega_{r}^{' } \cup \left \{ \mathbf{\hat{g}} _i \in \Omega_{l}^{' } \mid i=1,\dots , K -\left | \Omega_{r}^{' } \right |   \right \} 
\end{equation}
where $\left | \cdot  \right | $ indicates ther number of the set. In this manner, regardless of the number of valid candidates, the final sets $\Omega_{l}^{\ast }$ and $\Omega_{r}^{\ast }$ will each contain exactly $K$ proposals.

\textbf{Geometric Feature Extraction.} The angles between joints are critical to intuitively reflecting hand-related interactions. Based on this understanding, we extract joint geometric features from the left- and right-hand pose proposals. 
Formally, for the $i$-th proposal, the feature vector consisting of directional components of all non-repeated joint pairs is as follows:
\begin{equation}
\mathbf{f}  _i^{\tau } = \left [ {dx}_{jk}, {dy}_{jk} \mid \forall j,k \in \left \{ 1,\dots ,N_g \right \}, j < k   \right ] 
\end{equation}
where $\tau \in \left \{ l, r \right \} $ denotes the left-hand and right-hand proposals, respectively. And ${dx}_{jk} = \frac{\mathbf{\hat{g}}  _{i,2k-1} - \mathbf{\hat{g}}  _{i,2j-1}}{\left \| \mathbf{\hat{g}}  _{i,2k-1} - \mathbf{\hat{g}}  _{i,2j-1} \right \| } $ and ${dy}_{jk} = \frac{\mathbf{\hat{g}}  _{i,2k} - \mathbf{\hat{g}}  _{i,2j}}{\left \| \mathbf{\hat{g}}  _{i,2k} - \mathbf{\hat{g}}  _{i,2j} \right \| } $ are the normalized directional components on the $x$- and $y$-axis of the $j$-th and $k$-th joint pairs, respectively. By concatenating all the features of the left and right proposals, we obtain a global geometric vector, given by:
\begin{equation}
\mathbf{f} = \left [ \mathbf{f} _1^{l} ; \dots;  \mathbf{f} _K^{l} ;\mathbf{f} _1^{r} ; \dots;  \mathbf{f} _K^{r}    \right ] 
\end{equation}
$\mathbf{f}$ is a $2K N_g \left ( N_g - 1 \right ) $-dimensional vector, which not only captures rich inter-joint clues but also enhances our understanding of hand interactivity through gesture contexts from both hands.

\subsection{Interactivity Refinement}\label{sec4.3}
To obtain pose-aware interaction representations, we introduce hand pose prompts to refine the interaction-specific features using a pose-interaction attention mechanism. 

The refiner contains a self-attention layer \citep{carion2020end} that focuses on capturing and modeling the intrinsic correlations within the interaction features, obtaining the advanced interaction features ${\mathbf{I} }'= \left \{{\mathbf{I}_{i} }'\right \}^{N}_{i=1} $. 
Next, we introduce the pose offset-specific features $\mathbf{H} ^{\mathrm{off} }$ as pose prompts to inject pose awareness into the advanced interaction features. Specifically, we feed $\mathbf{H} ^{\mathrm{off} }$ into the attention mechanism as keys and values, while ${\mathbf{I} }' $ serves as queries. 
Each output element ${\mathbf{I}_i }'' $ is computed by aggregating all values weighted with attention: ${\mathbf{I}_i }''  =  {\textstyle \sum_{j}} \alpha_{ij}\left ( \mathbf{W}_v \mathbf{h} _j^{\mathrm{off} }\right ) $, where $\alpha_{ij}$ is the normalized attention weight, as follows:
\begin{equation}
\alpha _{ij} = softmax\left ( \frac{\left ( \mathbf{W}_q {\mathbf{I} }' _{i} \right )^{\mathrm{T} } \mathbf{W}_k\mathbf{h} _j^{\mathrm{off} } }{\sqrt{d} }  \right ) 
\end{equation}
where $\mathbf{W}_q $, $\mathbf{W}_k $, $\mathbf{W}_v $ are learnable embedding matrices corresponding to queries, keys, and values, respectively. After passing ${\mathbf{I} }' $ through the subsequent FFN layer, we finally obtain the refined pose-aware interaction representations $\mathbf{{I}^{\ast} } $. Our refiner contains only one decoder layer without consuming many computational resources. 
In this way, we guide the interaction features to focus on regions and features that are closely related to the subtle changes in hand poses.

\subsection{Feature Aggregation}\label{sec4.4}
To make the perception of interactivity more robust and effective, we aggregate the global hand geometric features $\mathbf{f} $ and the refined pose-aware interaction features $\mathbf{{I}^{\ast} } \in \mathbb{R}^{N \times d}$. 

First, the dimensions of both need to be aligned. To this end, we take a straightforward method: expand the dimensions of the feature vector $\mathbf{f}$ by repeating $N$ times. 
Next, we concatenate the tiled geometric feature map and the interaction features, and project them into a unified embedding space using an MLP. The feature aggregation can be formulated as follows:
\begin{equation}
\mathbf{E} = f_{emb} \left ( Concat\left [ \mathbf{I}^{\ast}, Tile\left ( \mathbf{f}  \right )  \right ]   \right ) 
\end{equation}

Using the enhanced embedding $\mathbf{E}$ as input to the interaction head significantly improves the model's performance compared to using only the refined interaction features $\mathbf{I}^{\ast} $. This improvement is attributed to the effective fusion of hand geometry and pose-aware interaction features, which complement each other and enhance the Ego-HOI detection model's reasoning about interactive behaviors.

\subsection{Training and Inference}
In addition to the hand bounding box and category prediction heads mentioned in Eq. \ref{eq: handhead}, our method employs another three heads to predict the verb category, object category, and object bounding box.

\textbf{Training Objective.} 
The baseline HOI detection methods are usually trained using a multi-task loss, as follows:
\begin{equation}
\mathcal{L}_{base} = \lambda _{L1} \mathcal{L} _{L1} + \lambda _{GIoU} \mathcal{L} _{GIoU} + \lambda _{hoc} \left ( \mathcal{L} _{oc} + \mathcal{L} _{hc} \right )   + \lambda _{ac} \mathcal{L} _{ac}
\label{eq:baseloss}
\end{equation}
where L1 loss \citep{ren2016faster} $\mathcal{L} _{L1} $ and GIoU loss \citep{rezatofighi2019generalized} $\mathcal{L} _{GIoU}$ are applied to both hand and object bounding box regression. To address the class imbalance problem in interaction classification, focal loss \citep{lin2017focal} $\mathcal{L} _{ac}$ is adopted. Notably, the Ego-HOI detection task requires explicit hand classification, which differs from the third-person perspective. Therefore, the cross-entropy loss is employed for both object classification $\mathcal{L} _{oc}$ and hand classification $\mathcal{L} _{hc}$. $\lambda _{L1}$, $\lambda _{GIoU}$, $\lambda _{hoc} $ and $\lambda _{ac} $ are hyper-parameters for weighting each loss. The loss functions of the baseline models \citep{tamura2021qpic, kim2023relational} and comparison models \citep{kim2021hotr, zhang2022exploring} are similar to Eq. \ref{eq:baseloss}, but the details may differ due to the unique characteristics of each model.

The learning of auxiliary hand pose estimation is supervised by the average L1 loss, as follows:
\begin{equation}
\mathcal{L} _{pose} =  \frac{1}{2N_g} \sum_{i=1}^{N} \sum_{j=1}^{2N_g} {\left | \mathbf{g}  _{i,j} - \mathbf{\hat{g}}  _{i,j} \right | }
\label{eq:poseloss}
\end{equation}
where $\mathbf{g} _{i,j}$ and $\mathbf{\hat g} _{i,j}$ are the ground truth and prediction result of the $j$-th value of the $i$-th hand pose candidate, respectively.

During training, the original loss $\mathcal{L}_{base} $ is integrated with the auxiliary pose estimation loss in Eq. \ref{eq:poseloss}. The overall loss $\mathcal{L} $ is given by:
\begin{equation}
\mathcal{L} = \mathcal{L} _{base} + \lambda_{pose} \mathcal{L} _{pose}
\end{equation}
where the weight $\lambda_{pose} $ denotes the weight to balance $\mathcal{L} _{base}$ and $\mathcal{L} _{pose}$, and is 1.0 by default.

\textbf{Inference.} 
Given a set of Ego-HOI predictions $\left \{ \left ( \mathbf{\hat{p} } _i^\mathrm{i} , \mathbf{\hat{p} } _i^\mathrm{h} , \mathbf{\hat{p} } _i^\mathrm{o} , \mathbf{\hat{b} } _i^\mathrm{h} , \mathbf{\hat{b} } _i^\mathrm{o}  \right )  \right \} _{i=1}^{N} $, where $\mathbf{\hat{p} } _{i}^{\mathrm{i} } \in \mathbb{R}^{\left | \mathcal{V} \right |  } $ and $\mathbf{\hat{p} } _{i}^{\mathrm{o} } \in \mathbb{R}^{\left | \mathcal{O} \right | +1 } $ correspond to the classification probabilities for interaction and object respectively, the predicted category $c_i^\tau $ and its score $s_i^\tau $ are given by 
$\underset{k}{\arg\max\,} \mathbf{\hat{p} }_{i,k}^\tau $ and $\underset{k}{\max\,} \mathbf{\hat{p} }_{i,k}^\tau $. 
Considering the hand classification, the confidence score of an Ego-HOI prediction is defined as follows:
\begin{equation}
s_{i}^{\mathrm{ehoi} } = s_{i}^{\mathrm{i} } \cdot  s_{i}^{\mathrm{h} } \cdot  s_{i}^{\mathrm{o} }
\end{equation}
We retain only the top predictions whose confidence scores exceed a predefined threshold from all $N$ predictions. Furthermore, to reduce false positives, we leverage the verb-object co-occurrence relationships shown in Fig.~\ref{fig3} to filter out implausible triplet combinations. This strategy ensures that the final Ego-HOI predictions are both confident and consistent with the realistic interaction patterns observed in the dataset.

\section{Experiments}

\subsection{Integration to Off-the-shelf HOI Detectors}
Our method is general and can be seamlessly integrated with most existing HOI detection approaches. The integration process is straightforward. In this work, we select two representative yet diverse baseline methods to evaluate the effectiveness of our proposed approach thoroughly.

\textbf{MUREN} \citep{kim2023relational} is an end-to-end Transformer-based approach with a three-branch architecture. It decouples human detection, object detection, and interaction classification, using independent decoder layers to extract task-specific tokens for sub-task learning. In our integration, the interaction branch's attention fusion module output is leveraged as the interaction representations $\mathbf{I} $, while the human branch's attention fusion module output serves as the hand representations $\mathbf{H} $.

\textbf{QPIC} \citep{tamura2021qpic} is one of the pioneering Transformer-based set prediction models for HOI detection. It employs a single decoder to predict all three elements of HOI: human, verb, and object. In our integration, the unified features output by the decoder are used indiscriminately as the original interaction features $\mathbf{I} $ and hand features $\mathbf{H} $. We apply a vanilla encoder to the unified features to derive the object-specific features.

\begin{table*}[t]
   \centering
   \caption{Performance and efficiency comparison of different representative HOI detection baselines before and after integrating our proposed HGIR scheme on the Ego-HOIBench dataset. All AP and Accuracy metrics are reported as percentages.}
   \label{tab3} 
  	\resizebox{1.0\linewidth}{!}{      

\begin{tabular}{llcccclcc}
\toprule
\multirow{2}{*}{Method} & \multirow{2}{*}{Backbone} & \multicolumn{4}{c}{Ego-HOIBench}                &                      & \multicolumn{2}{c}{Efficiency} \\
                        &                           & Full  $\uparrow$ & $\mathrm{mAP}_{50} \uparrow$  & $\mathrm{mAP}_{75} \uparrow$          &  Top@G Acc. $\uparrow$ &                      & \#Params $\downarrow$  & FPS $\uparrow$              \\ \hline
MUREN  \citep{kim2023relational}                 & ResNet-50                 & 65.0 & 83.5  &  75.6    & 81.4       &                      & 75M        & 15.9               \\ \rowcolor{gray!10}
+ \textit{HGIR (Ours)}                  & ResNet-50                 & 66.8\textcolor{red}{(+1.8)} & 84.1 & 76.3  & 85.7\textcolor{red}{(+4.3)}    &                      & 79M\textcolor{blue}{(+4M)}  & 15.0\textcolor{blue}{(-5.7\%)}       \\ 
QPIC \citep{tamura2021qpic}                   & ResNet-101                  & 76.7  & 84.8    & 80.8     & 86.5  & \multicolumn{1}{c}{} & 60M        & 18.6              \\ \rowcolor{gray!10}
+ \textit{HGIR (Ours)}                  & ResNet-101                & 78.4\textcolor{red}{(+1.7)}& 85.9 & 83.4  & 92.7\textcolor{red}{(+6.2)}    &                      & 65M\textcolor{blue}{(+5M)}   & 17.1\textcolor{blue}{(-8.1\%)}   \\
\bottomrule
\end{tabular}
}                 
\end{table*}

\subsection{Experimental Setup}
\textbf{Implementation Details.} Our experiments cover two baselines and their integrations with our method. We also include other existing HOI detection methods for comparison, all of which are modified and retrained for the Ego-HOI detection task. 
To obtain better detection performance, we fine-tune the object detector (usually DETR \citep{carion2020end} with a ResNet-50 backbone) on the Ego-HOIBench training set. 
All experiments are performed on 4 RTX 4090 GPUs. The hyper-parameters in the experiment remain consistent with the default settings of respective methods, but the batch size and initial learning rate are adjusted according to the supported computing resources. Specifically, all experiments of HOTR \citep{kim2021hotr} and MUREN \citep{kim2023relational} adopt a batch size of 16 and an initial learning rate of 5e-5. For STIP \citep{zhang2022exploring}, the HOI detector with a frozen object detector uses a batch size of 16 and an initial learning rate of 5e-5, while the batch size is 8 and the initial learning rate is 1e-5 during the two-stage joint fine-tuning. QPIC \citep{tamura2021qpic} is trained with a batch size of 8 and an initial learning rate of 5e-5.

\textbf{Evaluation Metrics.} We assess the models' performance on the Ego-HOIBench benchmark using mean average precision (mAP) with IoU thresholds ranging from 0.5 to 0.95 with a step size of 0.05. A detection result is considered a true positive only if the predicted hand, object, and verb categories are all correct, and the hand and object bounding boxes have IoUs with ground truths larger than the specified threshold. We further divide all the Ego-HOI triplet categories into rare and non-rare according to whether they appear at least 100 times in the training set. Based on this criterion, we report the mAP for the Full, Rare, and Non-rare categories. The mAPs of the full testing set at IoU thresholds of 0.5 and 0.75 are reported separately, denoted as $\mathrm{mAP_{50}} $ and $\mathrm{mAP_{75}} $, similar to \citep{zhang2023exploring, kim2023relational}. In addition, to highlight the improvement of our method in interaction recognition, we introduce Top@G Verb Accuracy as a metric. For an image to be considered correct, the G predictions with the highest probabilities must completely cover the set of true verb labels, where G represents the number of true labels.

\subsection{Improvement on Two Different Baselines}
Table \ref{tab3} shows the performance comparison of two mainstream baseline HOI detection methods before and after integrating our proposed method. By incorporating our approach, both baseline methods achieve significant performance improvements. Specifically, MUREN \citep{kim2023relational} achieves a 1.8\% improvement in Full mAP and a 4.3\% increase in Top@G Accuracy. As for QPIC \citep{tamura2021qpic}, Full mAP is improved by 1.7\% and Top@G Accuracy obtains a substantial improvement of 6.2\%, setting a new high for the state-of-the-art results. These results demonstrate that our method is applicable not only to models with a unified decoder but also to the methods that decouple the sub-tasks. Moreover, our scheme imposes no specific restrictions on the backbone. Note that after integrating our module, these two baseline methods can still maintain end-to-end training and reasoning.
We also compare their model sizes and runtime efficiencies to prove that the performance improvement is not due to the increase in model size. Although our method adds several million parameters, this increase is very limited relative to the original model size. Furthermore, in terms of Frames Per Second (FPS), the runtime speed drop is negligible, only a few percentage points. The results show that our technology is extremely lightweight and efficient.

\begin{table*}[t]
\centering
\caption{Detailed performance comparison of our proposed method (last row) and representative state-of-the-art HOI detection methods on the Ego-HOIBench dataset. \textit{Ours} adopts QPIC as a backbone followed by our proposed HGIR. All AP and Accuracy metrics are reported as percentages. $\dagger$ indicates joint fine-tuning of the object detector and HOI detector.}
\label{tab4}
\resizebox{1.0\linewidth}{!}{      
\begin{tabular}{llcccccccc}
\toprule
\multirow{2}{*}{Method} & \multirow{2}{*}{Backbone} & \multirow{2}{*}{End-to-End} & \multicolumn{5}{c}{Ego-HOI Detection}                                              &  & Interaction Recognition \\
                        &                           &                              & Full       & Rare       & Non-Rare & $\mathrm{mAP}_{50}$            & $\mathrm{mAP}_{75}$             &   & Top@G Accuracy                  \\ \hline
HOTR \citep{kim2021hotr}                   & ResNet-50                 & \checkmark                              & 22.4          & 9.8           & 27.3      & 44.3          & 18.9           &  & 85.1                          \\
STIP \citep{zhang2022exploring}                   & ResNet-50                 & $\times$                          & 25.0          & 10.8          & 30.4    & 47.7          & 22.4                 &  & 76.0                         \\
STIP$\dagger$ \citep{zhang2022exploring}                  & ResNet-50                 & $\times$                            & 32.1          & 15.9          & 38.4      & 57.9          & 30.6             &  & 78.4                      \\
MUREN \citep{kim2023relational}                 & ResNet-50                 & \checkmark                             & 65.0          & 61.6          & 66.3     & 83.5          & 75.6             &  & 81.4                          \\ 
QPIC \citep{tamura2021qpic}                  & ResNet-50                & \checkmark                           & 69.5 & 67.6              & 70.3  & 84.4              & 78.7    &           & 87.4                          \\
QPIC \citep{tamura2021qpic}                   & ResNet-101                & \checkmark                           & 76.7 & 75.5             & 77.1   & 84.8              & 80.8   &           & 86.5                           \\
\rowcolor{gray!10}
\textit{Ours}            & ResNet-101                       & \checkmark                          & \textbf{78.4} & \textbf{79.8} & \textbf{77.9}  & \textbf{85.9} & \textbf{83.4} &  & \textbf{92.7}                              \\ \bottomrule
\end{tabular}}
\end{table*}

\subsection{Comparison with State-of-the-art Methods}
Table \ref{tab4} presents a detailed performance comparison between our proposed method and many typical approaches, including the one-stage single-branch method QPIC \citep{tamura2021qpic}, the one-stage two-branch method HOTR \citep{kim2021hotr}, the one-stage three-branch method MUREN \citep{kim2023relational}, the two-stage method STIP \citep{zhang2022exploring} and its jointly fine-tuned version. Here, all AP and Accuracy metrics are presented in percentage form. We use QPIC as the baseline and integrate it with our scheme for comparison. Our method (last row) surpasses all existing one-stage and two-stage methods, whether in Ego-HOI detection or interaction recognition. A noteworthy phenomenon is that the rare triplet categories consistently underperform compared to the non-rare categories in terms of mAP across all other methods. In contrast, our method significantly enhances the detection performance of rare categories, even surpassing that of non-rare categories.
The superior performance of our method is mainly due to the fact that we effectively extract and incorporate hand pose cues into the interaction embedding. This enhancement significantly boosts the model's ability to distinguish complex and rare-seen interactions, further improving the overall performance of Ego-HOI detection.

\subsection{Ablation Study}
We conduct various ablation studies to validate the effectiveness of our method. For each ablation experiment, we modify one hyper-parameter or component while keeping all other hyper-parameters in their optimal settings. The MUREN baseline is used across all our ablation studies. We choose $\mathrm{mAP}_{50}$, Full mAP, and Top@G Accuracy as representative metrics to evaluate the performance of each variant.

\begin{table}[!t]
\caption{Ablation study evaluating the contribution of each component in our HGIR scheme. Starting from the baseline, components are added incrementally: Hand Pose Estimation (HPE), Interactivity Refinement (IR), and Hand Geometry Extraction (HGE). $\checkmark$ indicates the inclusion of the corresponding component.}
\label{tab5}
\centering
\begin{tabular}{cccccc}
\toprule
HPE & IR  & HGE & Full  & $\mathrm{mAP}_{50}$  & Top@G Acc. \\    \hline
-   & -   & -   & 65.0  & 83.5        & 81.4          \\
$\checkmark$   & -   & -        & 66.0 & 83.4       & 81.3        \\
$\checkmark$   & $\checkmark$ & -       & 66.1 & 83.6        & 84.7       \\ \rowcolor{gray!10}
$\checkmark$   & $\checkmark$  & $\checkmark$     & \textbf{66.8} &  \textbf{84.1}  &  \textbf{85.7}  \\\bottomrule
\end{tabular}
\end{table}

\textbf{Components of HGIR Scheme.}
To thoroughly assess the impact of each component in our method, we conduct an ablation study by gradually incorporating them into the baseline. The components evaluated include Hand Pose Estimation (HPE), Interactivity Refinement (IR), and Hand Geometry Extraction (HGE). The results are summarized in Table \ref{tab5}. 
Compared with the baseline, introducing a supervised HPE block results in a relative Full mAP gain of 1.0\%. This gain indicate that the auxiliary task enhances the learning of hand features, which indirectly positively impacts Ego-HOI detection. Next, integrating the IR block yields further advancements. While the gains in $\mathrm{mAP}_{50}$ and Full mAP are relatively modest, Top@G Accuracy achieves a significant leap to 84.7\%, with an increase of 3.4\%. These performance improvements show that incorporating pose prompts for engaging in meaningful interactions can significantly boost expressiveness.

Our complete method, as shown in the last row of Table \ref{tab5}, which includes the above two components and the PGE component, achieves notable improvements across all three metrics. Specifically, $\mathrm{mAP}_{50}$ is further increased by 0.5\%, and Full mAP is significantly improved by 0.7\%, and Top@G Accuracy by 1.0\%. These results demonstrate that the extracted hand geometric features provide complementary information, significantly enhancing interaction recognition and detection. The enhancements observed in this ablation study confirm the synergy of each component within the HGIR scheme and highlight the importance of utilizing hand geometric and refined interaction features to improve the model's accuracy and robustness in Ego-HOI perception.

\textbf{Pose Estimation Schemes.} We compare the impact of different pose estimation schemes, as shown in Table \ref{tab6}. We explore two main categories of methods: directly predicting hand joint positions from the hand features and indirectly estimating them by combining reference points and offsets. When directly predicting (row a), we observe that both $\mathrm{mAP}_{50}$ and Full mAP are the lowest among the four schemes. The challenge with this scheme is that it is equivalent to predicting the offsets using the upper left corner of the image as a reference point. The long distance between the reference point and the hand makes accurate prediction extremely difficult.

\begin{table}[!t]
\caption{Ablation study comparing different hand pose estimation schemes. ``Direct'' predicts joint positions directly; ``Learnable Point'', ``Center Point'', and ``Top-Center Point'' use different reference points for offset-based estimation.}
\label{tab6}
\centering
\begin{tabular}{ccccc}
\toprule
& Method           & Full  & $\mathrm{mAP}_{50}$  & Top@G Acc.  \\ \hline
(a) & Direct  & 65.7   & 82.9  &  85.3     \\
(b) & Learnable Point    & 66.6 & 83.1       & 85.3        \\  
(c) & Center Point       & 66.5  & 83.6      & 84.3        \\
\rowcolor{gray!10}
(d) & Top-Center Point & \textbf{66.8} & \textbf{84.1} & \textbf{85.7} \\ \bottomrule\\
\end{tabular}
\end{table}

Various schemes for computing reference points are evaluated, ranging from learnable points to hand box centers and top centers. Compared to direct prediction, leveraging hand-detection-specific features to infer reference points (row b) significantly improves Full mAP by 0.9\%. However, the notable improvement in Full mAP is not synchronously reflected in the other two metrics. 
In contrast, using the centers (row c) or top centers (row d) of the predicted hand boxes as references achieve better results in terms of $\mathrm{mAP}_{50}$. The best performance is achieved using the top center reference points, with $\mathrm{mAP}_{50}$ increased to 84.1\%, Full mAP increased to 66.8\%, and Top@G Accuracy reaching 85.7\%. These improvements are likely due to our ability to explicitly constrain the reference points and estimated joint positions to the vicinity of the hand, leading to more stable and accurate joint localization and further enhancing the overall Ego-HOI detection performance.

\textbf{Number of Selected Pose Proposal Pairs.} We also study the impact of the number of selected pose proposal pairs on model performance. Specifically, we test different values of $K$ (1, 2, 3, and 4), where $K$ represents that only the top $K$ pairs of left-hand and right-hand pose proposals with highest scores are used to extract hand geometric features. The results are summarized in Table \ref{tab7}. Our observations show that the model performs best when $K=1$. 
We speculate that increasing the number of proposal pairs may introduce more invalid or low-quality geometric features, which dilutes the effective information and negatively impacts the stability of relational reasoning.

\begin{figure*}[!t]
\centering
\includegraphics[width=6.4in]{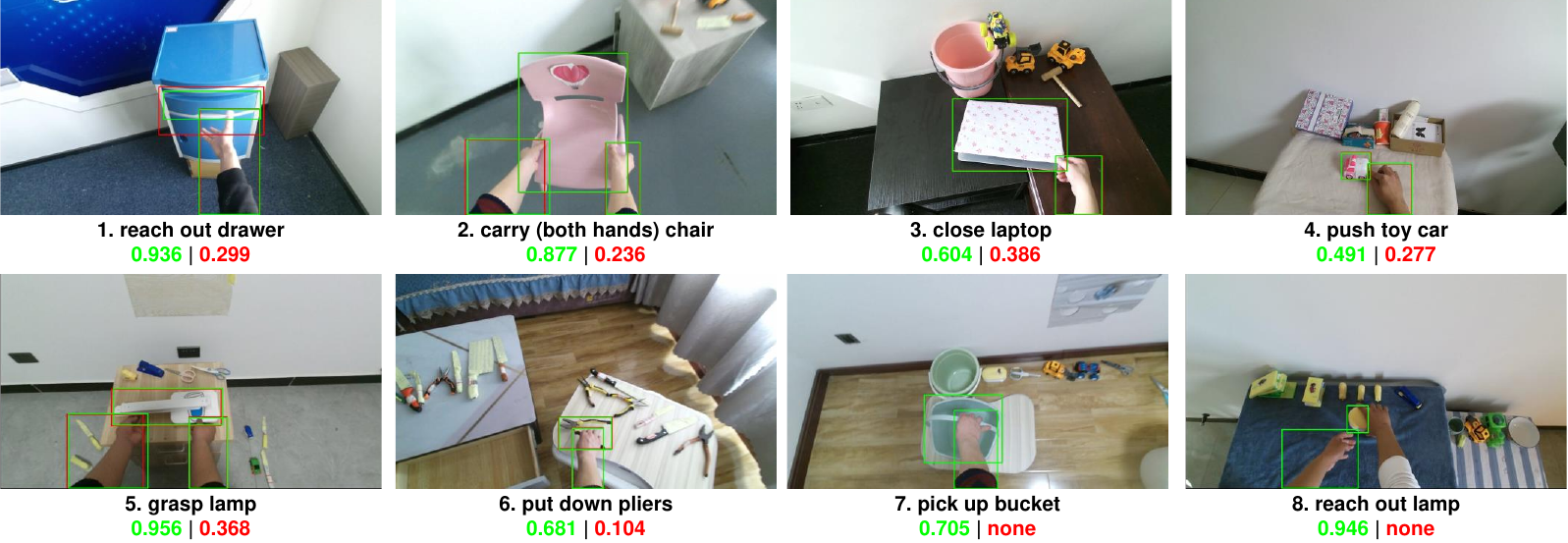}
\caption{Qualitative comparison between the baseline and our proposed method on challenging Ego-HOI detection cases. For each image, predictions from our method are shown in \textcolor{green}{green}, and baseline predictions in \textcolor{red}{red}. The predicted triplet classes and confidence scores are shown below each image. If the model fails to output a true positive, the score is marked as \textbf{none}. For clarity, hand categories are omitted from the captions.}
\label{fig6}
\end{figure*}

\begin{table}[!t]
\caption{Performance comparison of different number of selected pose proposal pairs.}
\label{tab7}
\centering
\begin{tabular}{cccc}
\toprule
\begin{tabular}[c]{@{}c@{}}\# Selected Pose\\ Proposal Pairs ($K$)\end{tabular} & Full &  $\mathrm{mAP}_{50}$ & Top@G Acc. \\ \hline \rowcolor{gray!10}
1 & \textbf{66.8} & \textbf{84.1}  & \textbf{85.7}    \\
2                                                          & 66.0 & 83.6  & 85.7     \\
3                                                                        & 65.4 & 82.5 & 82.0 \\ 
4                                                                         &  66.4 & 83.6 & 81.8 \\ \bottomrule
\end{tabular}
\end{table}

\subsection{Qualitative Results and Discussions}

To qualitatively demonstrate the advantages of our method in Ego-HOI detection, comparison examples between the baseline and our proposed method are provided in Fig. \ref{fig6}. Our method is particularly outstanding in improving the confidence of interaction predictions. For instance, in Case 1, the baseline model predicts a \textit{right-hand reach out drawer} with a score of 0.299, while our model significantly improves this score to 0.936. Furthermore, our method successfully recognizes Ego-HOI triplets that the baseline method fails to output a true positive prediction (Cases 7 and 8). These improvements cover scenes with small or occluded objects (Samples 4, 6, 7, 8) and complex scenes (Samples 2, 5, 8), showcasing that our approach can provide more accurate predictions under challenging conditions. Overall, our method shows apparent advantages in prediction accuracy and robustness.

\begin{figure*}[!t]
\centering
\includegraphics[width=3.5in]{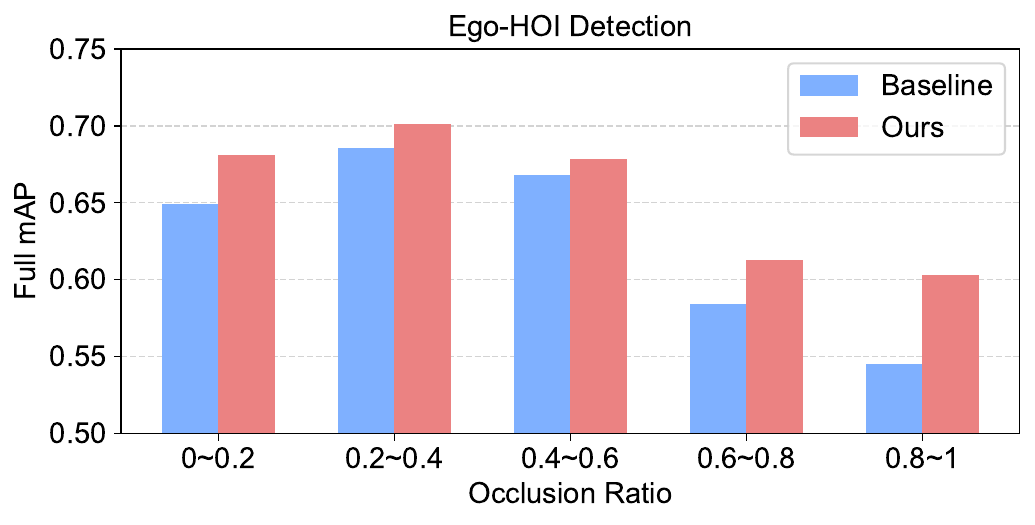}
\includegraphics[width=3.5in]{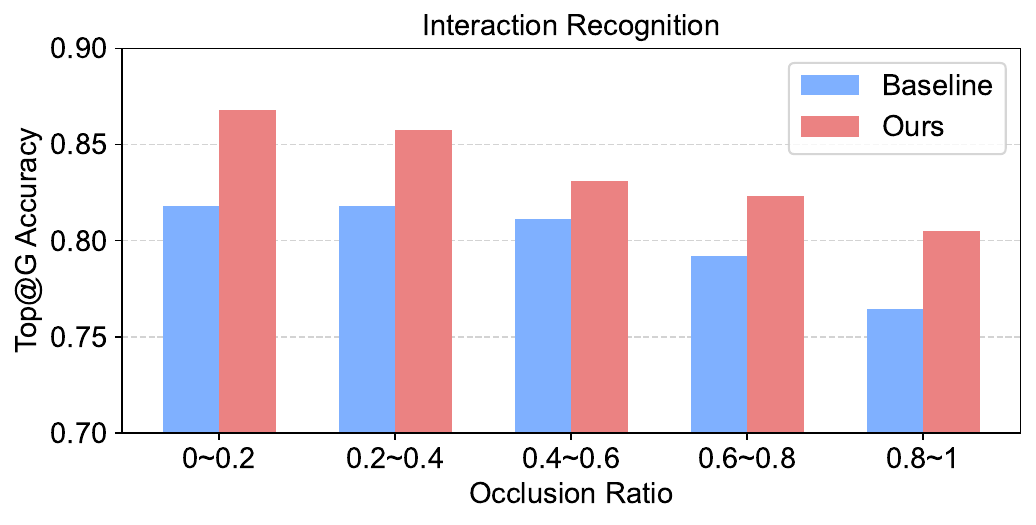}
\caption{Comparison of our method and the baseline on Ego-HOI detection (top) and interaction recognition (bottom) across different object occlusion ratios. }
\label{fig7}
\end{figure*}

We also compare our proposed method with the baseline across different object occlusion ratios. For Ego-HOI detection, we count prediction results according to their ground truth occlusion ratios, as shown in Fig. \ref{fig7} (top). For interaction recognition, we statistically classify each prediction according to the average ground truth occlusion ratio of the instances within each image, as shown in Fig. \ref{fig7} (bottom). Overall, the performance of both the baseline and our method shows a downward trend with the increase of occlusion ratio. This phenomenon occurs because occlusions may obscure critical features, thus hindering the model's learning. Nonetheless, our method consistently outperforms the baseline at all occlusion levels. In particular, at the high occlusion level (0.8$\sim$1), our method improves Full mAP by 5.7\% and Top@G Accuracy by 4.0\% compared to the baseline. These significant improvements are mainly due to our method's ability to leverage poses as additional cues to enhance interaction features and infer interactions more effectively, even when the visible portion of an object is too limited to provide enough information.

\section{Conclusion}
In this paper, we introduce a new benchmark dataset, Ego-HOIBench, which aims to advance the research of human-object interaction detection from an egocentric perspective. Ego-HOIBench provides high-quality, fine-grained annotations of hand-verb-object triplets, covering a diverse range of scenarios, object types, and hand configurations. To address the unique challenges of Ego-HOI detection, particularly severe occlusions and complex hand-object interactions, we propose the Hand Geometry and Interactivity Refinement (HGIR) scheme. HGIR leverages global hand pose cues and pose-aware interaction features to enhance interaction representation in a lightweight and plug-and-play manner. Extensive experiments demonstrate that our method can be seamlessly integrated into various HOI detection frameworks, consistently improving both detection and interaction recognition performance on Ego-HOIBench. 

Future work will focus on addressing the imbalanced issue and further improving model generalization to rare and complex Ego-HOI triplets. We hope Ego-HOIBench and our proposed method will facilitate further research and practical advancements in egocentric vision and human-object interaction understanding.

\section*{Acknowledgments}
The research work was conducted in the JC STEM Lab of Machine Learning and Computer Vision funded by The Hong Kong Jockey Club Charities Trust.

\printcredits

\bibliographystyle{model5-names}

\bibliography{cas-refs}


\end{document}